\documentclass[letterpaper, 10pt, conference]{ieeeconf}
\IEEEoverridecommandlockouts                              
\usepackage{amsmath}    
\usepackage{amsfonts}
\usepackage{graphicx}   

\usepackage{cite}

\usepackage{float}
\usepackage{afterpage}
\usepackage{algorithmic}
\usepackage{algorithm}

\usepackage{array}
\usepackage{graphicx}   
\usepackage{url}

\usepackage{pgf,tikz}
\usepackage{mathrsfs}
\usetikzlibrary{arrows}
\usepackage{subcaption}

\usepackage{url}
\usepackage{array}
\newcolumntype{P}[1]{>{\centering\arraybackslash}p{#1}}

\usepackage{color}
\usepackage{comment}

\newcommand{\R}{\mathbb R}

\newcommand{\cset}{\mathcal U}

\newcommand{\state}{x}
\newcommand{\targetset}{\mathcal L}
\newcommand{\ctrl}{u}
\newcommand{\dyn}{f}
\newcommand{\traj}{\xi}
\newcommand{\idyn}{g}
\newcommand{\fset}{\mathbb F}
\newcommand{\cfset}{\mathbb U}
\newcommand{\ctime}{\tau}
\newcommand{\rect}{\psi}
\newcommand{\weights}{\mathbb W}
\newcommand{\keep}{A}
\newcommand{\dataset}{\mathcal D}
\newcommand{\trainset}{\mathcal X}
\newcommand{\emax}{M_\epsilon} 
\newcommand{\controlset}{\mathcal V}

\newcommand{\costset}{\mathcal X_C}

\newcommand{\pos}{p}

\title{\LARGE Using Neural Networks to Compute Approximate and Guaranteed Feasible
Hamilton-Jacobi-Bellman PDE Solutions}

\author{Frank Jiang*, Glen Chou*, Mo Chen*, Claire J. Tomlin
\thanks{This research is supported by ONR under the Embedded Humans MURI (N00014-16-1-2206).}
\thanks{* These authors contributed equally to this work. All authors are with the Department of Electrical Engineering and Computer Sciences, University of California, Berkeley. \{fjiang6o2,gchou,mochen72,tomlin\}@berkeley.edu}
}

\begin{document}

\maketitle

\thispagestyle{empty}
\pagestyle{empty}


\begin{abstract}
To sidestep the curse of dimensionality when computing solutions to
Hamilton-Jacobi-Bellman partial differential equations (HJB PDE), we propose an
algorithm that leverages a neural network to approximate the value function. We
show that our final approximation of the value function generates near optimal
controls which are guaranteed to successfully drive the system to a target
state. Our framework is not dependent on state space discretization, leading to
a significant reduction in computation time and space complexity in comparison
with dynamic programming-based approaches. Using this grid-free approach also
enables us to plan over longer time horizons with relatively little additional
computation overhead. Unlike many previous neural network HJB PDE approximating
formulations, our approximation is strictly conservative and hence any
trajectories we generate will be strictly feasible. For demonstration, we
specialize our new general framework to the Dubins car model and discuss how the framework can be applied to other models with higher-dimensional state
spaces.
\end{abstract}



\section{Introduction}

In recent years, rapid progress in robotics and artificial intelligence has accelerated the need for efficient path-planning algorithms in high-dimensional spaces. In particular, there has been vast interest in the development of autonomous cars and unmanned aerial vehicles (UAVs) for civilian purposes \cite{JPDO13, Fehrenbacher2016, Amazon16, BBC16, AUVSI16, Kopardekar16}. As such systems grow in complexity, development of algorithms that can tractably control them in high-dimensional state spaces are becoming necessary.



Many path planning problems can be cast as optimal control problems with initial and final state constraints. Dynamic programming-based methods for optimal control recursively compute controls using the Hamilton-Jacobi-Bellman (HJB) partial differential equation (PDE). Such methods suffer from space and time complexities that scale exponentially with the system dimension. Dynamic programming is also the backbone for Hamilton-Jacobi (HJ) reachability analysis, which solves a specific type of optimal control problem and is a theoretically important and practically powerful tool for analyzing a large range of systems. It has been extensively studied in \cite{Varaiya67, Evans84, Barron90, Tomlin00, Mitchell05, Margellos11, Bokanowski10, Fisac15}, and successfully applied to many low-dimensional real world systems \cite{Ding08, Chen15b, Mitchell05}. 
We use the reachability framework to validate our method.

To alleviate the curse of dimensionality, many proposed dynamic programming-based methods heavily restrict the system at hand \cite{Majumdar13, Dreossi16,Darbon16}. Other less restrictive methods use projections, approximate dynamic programming, and approximate systems decoupling \cite{Mitchell03, McGrew08, Chen16b}, each limited in flexibility, scalability, and degree of conservatism. There are also several approaches towards scalable verification \cite{SpaceEx, C2E2, Flow_Star, dReach}; however, to the best of our knowledge, these methods do not extend easily to control synthesis. Direct and indirect methods for optimal control, such as nonlinear model predictive control \cite{Allgower04}, the calculus of variations \cite{Gelfand63}, and shooting methods \cite{Aronna13}, avoid dynamic programming altogether but bring about other issues such as nonlinearity, instability, and sensitivity to initial conditions. 


One primary drawback of dynamic programming-based approaches is the need to compute a value function over a large portion of the state space. This is computationally wasteful since the value function, from which the optimal controller is derived, is only needed along a trajectory from the system's initial state to the target set. A more efficient approach would be to only compute the value function local to the trajectory from the initial state to the target set. However, there is no way of knowing where such a trajectory will lie before thoroughly computing the value function.

Methods that exploit machine learning have great potential because they are state discretization-free and do not depend on the dynamic programming principle. Unfortunately, many machine learning techniques cannot make the guarantees provided by reachability analysis. For instance, \cite{Becerikli2003} and \cite{Kim1997} use neural networks (NNs) as nonlinear optimizers to synthesize trajectories which may not be dynamically feasible. The authors in \cite{Allen2014} propose a supervised learning-based algorithm that depends heavily on feature tuning and design, making its application to high-dimensional problems cumbersome. In \cite{Djeridane2006} and \cite{Niarchos2006}, the authors successfully use NNs for approximating the value function, but the approximation is not guaranteed to be conservative. 

In this paper, we attempt to combine the best features of both dynamic
programming-based optimal control and machine learning using an NN-based
algorithm. Our proposed grid-free method conservatively approximates the value
function in only a region around a feasible trajectory. Unlike previous machine learning techniques, our technique guarantees a direction of conservatism, and unlike previous dynamic programming-based methods, our approach involves an NN that effectively finds the relevant region that requires a value function. Our contributions will be presented as follows:

\begin{itemize}
\item In Section \ref{sec:prob_form}, we summarize optimal control and the formalisms used for this work.
\item In Section~\ref{sec:overview}, we provide an overview of the full method and the core ideas behind each stage, as well as highlight the conservative guarantees of the method.
\item In Sections \ref{sec:training_phase} and \ref{sec:post}, we present the
two phases of our proposed algorithm and the underlying design choices.
\item In Section \ref{sec:dubins_ex}, we illustrate our guaranteed-conservative approximation of the value function, and the resulting near-optimal trajectories for the Dubins Car.
\item In Section \ref{sec:conclusion}, we conclude and discuss future directions.
\end{itemize}


\section{Problem Formulation}\label{sec:prob_form}
In this section, we will provide some definitions essential to express our main results. Afterwards, we will briefly discuss the goals of this paper.

\subsection{Optimal Control Problem}
Consider a dynamical system governed by the following ordinary differential equation (ODE):

\begin{equation}
\label{eq:dyn}
\begin{aligned} 
    \dot \state = \dyn(\state, \ctrl), t \in [t_0, 0] \\
    \state(t_0) = \bar \state, \quad \ctrl \in \cset
\end{aligned}
\end{equation}

Note that since the system dynamics \eqref{eq:dyn} is time-invariant, we assume without loss of generality that the final time is $0$.

Here, $\state \in \R^n$ is the state of the system and the control function $\ctrl(\cdot)$ is assumed to be drawn from the set of measurable functions. Let us further assume that the system dynamics $\dyn: \R^n \times \cset \rightarrow \R^n$ are uniformly continuous, bounded, and Lipschitz continuous in $\state$ for fixed $\ctrl$. Denote the function space from which $\dyn$ is drawn as $\fset$.

With these assumptions, given some initial state $\state$, initial time $t_0$, and control function $\ctrl(\cdot) \in \cfset$, there exists a unique trajectory solving \eqref{eq:dyn}. We refer to trajectories of (\ref{eq:dyn}) starting from state $\state_1$ and time $t_1$ as $\traj^{\dyn}(t; \state_1, t_1,\ctrl(\cdot))$, with $\state_1\in\R^n$ and $t_0 \le t, t_1 \le 0$. Trajectories satisfy an initial condition and \eqref{eq:dyn} almost everywhere:

\begin{equation}
\label{eq:traj}
\begin{aligned}
\frac{d}{dt} \traj^\dyn(t; \state_1, t_1, \ctrl(\cdot)) = \dyn(\traj^\dyn(t; \state_1, t_1, \ctrl(\cdot)), \ctrl(t)) \\
\traj^\dyn(t_1; \state_1, t_1, \ctrl(\cdot)) = \state_1
\end{aligned}
\end{equation}

Note that we can use the trajectory notation to specify states that satisfy a final condition if $t \le t_1$. In this paper this will often be the case, since our NN will be producing backward-time trajectories.

Consider the following optimal control problem with final state constraint\footnote{For simplicity we constrain the final state to a single state; our method easily extends to the case with a set-based final state constraint, $\state(0) \in \targetset$.}:

\begin{equation}
\label{eq:value_func}
\begin{aligned}
V(\state, t) = \min_{\ctrl(\cdot), t_0} \int_t^0 C(\ctrl(\tau)) d\tau \\
\text{subject to } \state(t_0) = \bar\state, \state(0) = \state_\targetset
\end{aligned}
\end{equation}

The value function $V(\state, t)$ is typically obtained via dynamic programming-based approaches such as \cite{Barron89, Mitchell05, Bokanowski10, Fisac15, Yang2013, Sethian96, Osher02, Mitchell07}, and an appropriate HJB partial differential equation is solved backwards in time on a grid representing the discretization of states. Once $V(\state, t)$ is found, the optimal control, which we denote $\ctrl^*(\state, t)$, can be computed based on the gradient of $V(\state, t)$:

\begin{equation}
\label{eq:opt_control}
\ctrl^*(\state, t) = \arg \min_{\ctrl \in \cset} \nabla V(\state, t) \cdot \dyn(\state, \ctrl)
\end{equation}

Unfortunately, the computational complexity of these methods scales exponentially with the state space dimension.

\subsection{Goal}
In this paper, we seek to overcome the exponentially scaling computational complexity. Our approach is inspired by two inherent challenges that dynamic programming-based methods face. First, since only relatively mild assumptions are placed on the system dynamics \eqref{eq:dyn}, optimal trajectories are \textit{a priori} unknown and could essentially trace out any arbitrary path in the state space. Dynamic programming ignores this issue by considering \textit{all possible} trajectories. Second, in a practical setting, the system starts at some particular state $\state_0$. Thus, the optimal control, and in particular $\nabla V(\state, t)$ in \eqref{eq:opt_control}, is needed only in a ``corridor'' along the optimal trajectory. However, since the optimal trajectory is \textit{a priori} unknown, dynamic programming-based approaches resort to computing $V(\state, t)$ over a very large portion of the state space so that the $\nabla V(\state, t)$ is available regardless of where the optimal trajectory happens to be.

In this paper, we propose a method that, in contrast to dynamic programming-based methods,

\begin{enumerate}
\item\label{item:fast} has a substantially faster computation time and smaller memory-usage;
\item\label{item:general} is a flexible and general framework that can be applied to higher-order systems with just hyperparameter tuning; 
\item\label{item:small} generates an approximate value function $\hat V(\state,t)$ from which a controller that drives the system to the target can be synthesized;
\item\label{item:conservative} guarantees that $\hat V(\state,t) \ge V(\state, t) ~ \forall x, t$, so that a direction of conservatism can be maintained despite the use of an NN.
\end{enumerate}

We enforce \ref{item:fast}) by avoiding operations that exhaustively search the state space. As seen in Section~\ref{subsec:dynamic}, the training and final data sets are either generated randomly or outputted by the NN, both constant time operations. Furthermore, we rely on NNs being universal function approximators \cite{Cybenko1989} to make our method general to the system dynamics, thus satisfying~\ref{item:general}). While our method does have some limitations, as discussed at the end of Section \ref{subsec:nnet}, we do not believe that these limitations are restrictive in the context of optimal control. Our post-processing of the NN outputs outlined in Section \ref{sec:post} satisfies \ref{item:small}). Finally, we use the dynamics of the system to ensure our final output satisfies \ref{item:conservative}); this is detailed in Section \ref{subsec:dynamic}.

Our method overcomes the challenges faced by dynamic programming-based methods
in two phases: the NN training phase, which allows the NN to learn the inverse
backward system dynamics while also generating a dataset from which we can
obtain a conservative approximation of the value function; and the controller
synthesis phase, which uses the approximation to synthesize a controller to
drive the system from its initial state to its target.



\section{Overview}\label{sec:overview}

\begin{figure}[!t]
    \centering
    \includegraphics[width=0.9\columnwidth]{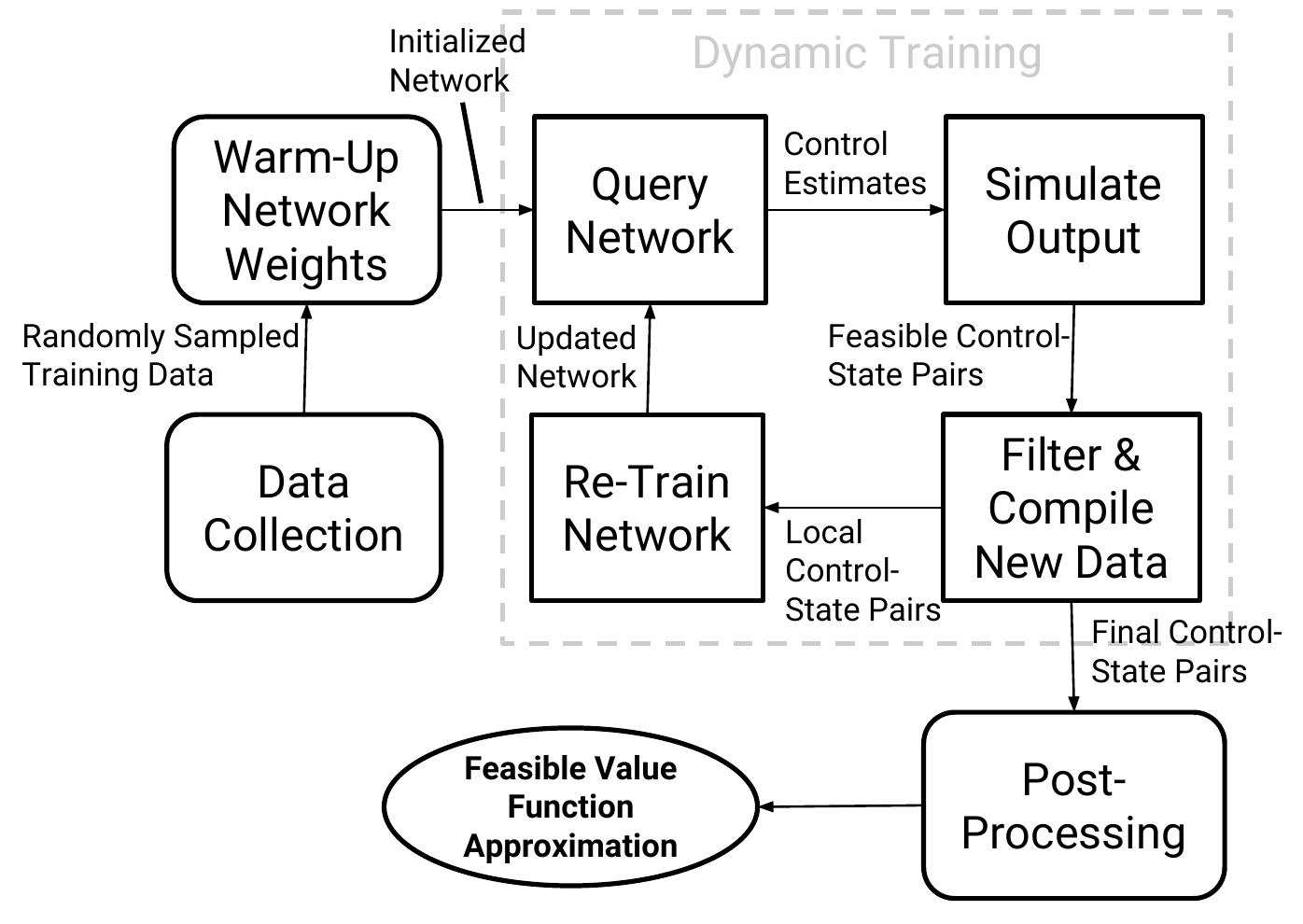}
    \caption{Stages of proposed method}
    \label{fig:overview}
\end{figure}

In this section we will briefly describe the different stages of our computational
framework (depicted in Fig.~\ref{fig:overview}). We leave the details of each
stage to subsections of Section~\ref{sec:training_phase} and \ref{sec:post}.


\subsection{Pre-Training}\label{subsec:pre_train}
In the pre-training phase (detailed in Section~\ref{subsubsec:warmup}), we
produce some initial datasets to prepare the NN for the dynamic training
procedure. The dynamic training procedure is already designed to help the NN
improve its performance iteratively; however, the pre-training phase speeds up
the convergence by warming up the NN with sampled data of the system dynamics, $\dyn$.

\subsection{Dynamic Training}\label{subsec:dyn_train}
Our proposed dynamic training loop is tasked with both improving the approximation
of the NN and producing the necessary data from which we extract our value
function approximation. Fortunately, these two tasks are inherently tied to
one another.

We pose our NN as an approximation of the inverse dynamics of the
system (this is formally clarified in Section~\ref{sec:training_phase}). In
other words, if we query our NN with a set of states to check its
approximation, then the NN will predict a set of controls to drive the system
to those states (detailed in Section~\ref{subsubsec:query}). We can check the
correctness of this prediction by simulating the controls through $\dyn$
(detailed in Section~\ref{subsubsec:sim}). Even if this prediction is
inaccurate, we can recycle the new data the NN produced through an
attempted prediction, and add the data into our training set to re-train the
NN. Since we simulated the NN's prediction to get the \textit{actual} states the
predicted control drives the system to, we now have corrective data with which
we can re-train our NN. 
By doing this repeatedly, we iteratively train the NN with feedback.
The data produced by the NN can be used to evaluate an approximate value of the value function using (\ref{eq:value_func}). Thus, with this dynamic
training loop, we are able to produce data to construct an approximate value
function while iteratively improving our NN's prediction capabilities.
Furthermore, as the NN improves, the relevance of the data also improves.

To further encourage the process, in every iteration we additionally apply
stochastic filters to our dataset that favor more local and optimal data
(detailed in Section~\ref{subsubsec:nnet_filt}). This way, we can ensure the
NN will not saturate or keep any unnecessary data for our final value
function approximation.

\subsection{Post-Training}\label{subsec:post_train}
Once dynamic training is complete, the NN 
will be able to make accurate predictions and our dataset will
encompass the region over which an approximate value function is needed. After
some post-processing over the dataset (detailed in
Section~\ref{subsec:post_process}), we will show that we can successfully
extract a value function approximation from which we synthesize control to drive
our system from our initial state to our goal state.

\section{Neural Network Training Phase} \label{sec:training_phase}

Given a target state $\state_\targetset$ and assuming the system starts at some initial state $\bar \state$, we want to design an NN that can produce a control function $\ctrl(\cdot)$ that drives the backward-time system from $\state_\targetset$ to $\bar\state$. More concretely, consider the inverse backward dynamics of the system, denoted
$\idyn_{{\state_{\targetset}}}: \R^n \times \fset \rightarrow \cfset$, and defined to be

\begin{equation}
\label{eq:inverse_sys}
\idyn_{\state_\targetset}(\bar\state; -\dyn(\cdot, \cdot)) = \ctrl^*(\cdot) \\
\end{equation}

\noindent where the optimal control $\ctrl^*(\cdot)$ is defined for $t \le 0$. For simplicity, we will write $\idyn_{\state_\targetset}$
as $\idyn$ from now on. Given some $\ctrl^*(\cdot)$ defined in the time interval $[t, 0]$, we have $\bar\state = \traj^{\dyn}(0; \state_\targetset, t, \ctrl^*(\cdot))$.

Our NN is an approximation of $\idyn$, and we will denote the NN as $\hat\idyn$.
Let $\hat\ctrl(\cdot)$ be the control produced by $\hat\idyn$, and let the time
interval for which $\hat\ctrl(\cdot)$ is defined be denoted $[-\hat T, 0]$. The
primary tasks of our training procedure will be to:
\begin{enumerate}
    \item iteratively improve our NN's training set so $\hat\idyn$ approaches
        $\idyn$ in the region local to the path given by $\traj^{\dyn}(0;
        \state_\targetset, t, \ctrl^*(\cdot)), t \le 0$; and
    \item produce a dataset of states, their corresponding control and, in turn,
        the corresponding value approximation, which will be used for control
        synthesis.
\end{enumerate}


\subsection{Neural Network Architecture} \label{subsec:nnet}

\begin{figure}
\centering
\includegraphics[width=0.9\columnwidth]{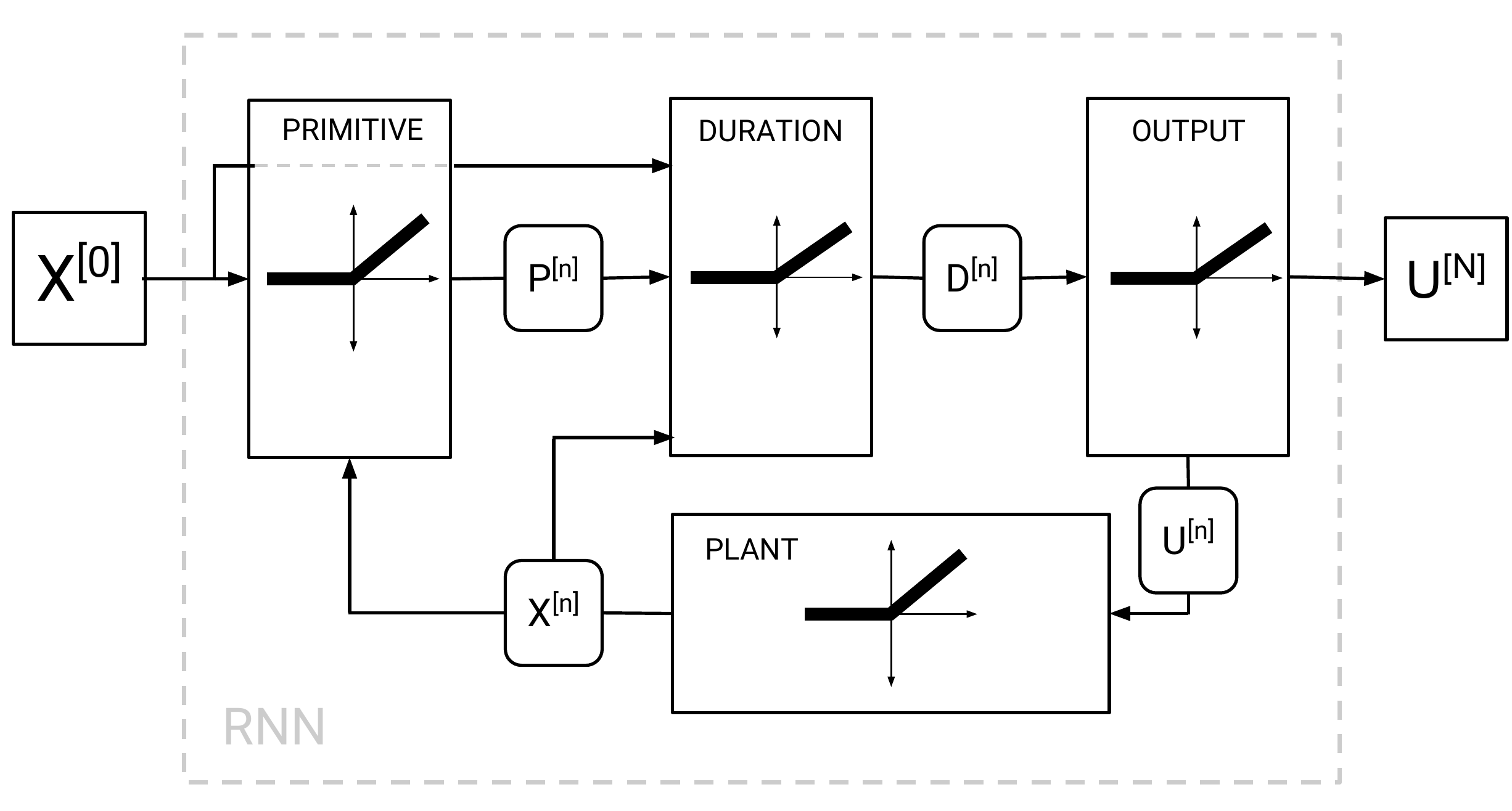}
\caption{Four-layer RNN which takes input states and outputs discretized control.}
\label{fig:nnet_architecture}
\end{figure}

Denote the maximum time horizon as $\bar T$. To reduce the space in which the NN needs to look for candidate control functions, we assume that $\hat \ctrl(\cdot)$ returned by the NN is composed of two finite sequences $\{\ctrl^j\}_{j=1}^K, \{\ctime^j\}_{j=1}^K$ called the sequence of control primitives and the sequence of time durations respectively. Mathematically, the control function $\hat \ctrl(\cdot)$ is of the form

\begin{equation}
\label{eq:u_recon}
\hat\ctrl(t) =
\begin{cases}
\ctrl^1, &\quad t \in [-\sum_{j=1}^K\ctime^j, -\sum_{j=2}^K\ctime^j] \\
\ctrl^2, &\quad t \in (-\sum_{j=2}^K\ctime^j, -\sum_{j=3}^K\ctime^j] \\
\cdots &\\
\ctrl^{K-1}, &\quad t \in \left(-\ctime^K-\ctime^{K-1}, -\ctime^K\right] \\ 
\ctrl^K, &\quad t \in \left(-\ctime^K, 0\right] \\ 
\end{cases}
\end{equation}

\noindent where we implicitly define $\hat T = \sum_{j=1}^K\ctime^j$. Since we will only use the control to obtain the approximation $\hat V(\cdot, \cdot)$ and not for actually controlling the system, we do not need the generated controls to be extremely accurate or continuous, as we will explain in Sections \ref{subsec:dynamic} and \ref{sec:post}.

We propose a rectified linear unit recurrent NN (RNN) with the following
structure:

\begin{equation*}
\begin{aligned}
\text{Primitive Layer: } &P^{[n]} = \rect(W_P \cdot X^{[n]} + b_P) \\
\text{Duration Layer: } &D^{[n]} = \rect(W_{D_1}\cdot P^{[n]} + W_{D_2} \cdot X^{[n]} + b_D) \\
\text{Control Layer: } &U^{[n]} = W_L\cdot D^{[n]} + b_L \\
\text{Plant Layer: } &X^{[n+1]} = \rect(W_X\cdot U^{[n]} + b_X)
\end{aligned}
\end{equation*}

\noindent where $\rect$ is the positive rectifying function defined as $\rect(a)
= \max(0, a)$ and $n = 0, 1, \ldots, N$. The input of the RNN is $\bar \state$
($X^{[0]} = \bar \state$), and the output is some $\hat \ctrl(\cdot)$ ($U^{[N]}
= \hat\ctrl(\cdot)$), that approximately brings the system from $\bar \state$ to
$\state_\targetset$. The parameters we learn through training are the weights
$W_P$, $W_{D_1}$, $W_{D_2}$, $W_L$, $W_X$ and biases $b_P, b_D, b_L, b_X$. All
weights and biases will be collectively denoted $\weights$. For prediction,
$\hat u(\cdot)$ and training example, $u(\cdot)$, we discretize both controls
using (\ref{eq:u_recon}), then the training is
performed with mean-squared error (MSE), or
\begin{equation}
\label{eq:MSE}
MSE = \frac 1n \sum_{n=1}^{N} (\hat\ctrl^n - \ctrl^n)^2
\end{equation}
as the cost function, where $\ctrl(\cdot)$ is our training example output.



As already mentioned, the primitive layer takes a state as input and computes a
control primitive. The duration layer takes in the primitive layer's output and
the same input state, and outputs a time duration. This time duration is then
passed through the control (also called output) linear layer, which outputs the
sequences $\{\ctrl^j\}, \{\ctime^j\}$ representing the control function $\hat
\ctrl(\cdot)$. Afterwards, the control function is fed into the plant layer,
which attempts to encode the backward dynamics $-\dyn$. The plant's output state, $X^{[n+1]}$, is then fed back to the primitive layer. 

Explicitly, the RNN can be written as

\begin{equation}
  \label{eq:RNN}
  \hat \idyn(\bar \state, \state_\targetset; -\dyn(\cdot,\cdot), \weights) = \hat \ctrl(\cdot) = U^{[N]}
\end{equation}

\noindent where $\hat\ctrl(\cdot)$ is given in the form of the sequences $\{\ctrl^j\}, \{\ctime^j\}$.

In the next section, we discuss the dynamic training procedure for this RNN.

\subsection{Detailed Dynamic Training}\label{subsec:dynamic}

\subsubsection{Warm-up and Initial Training} \label{subsubsec:warmup}

We first train the RNN without knowledge of $\bar\state$. For starters, we will require
training examples in the form of $\{\hat\state_i, \hat\ctrl_i(\cdot)\}$ that
sufficiently capture the basic behaviors of the system dynamics. To do this, we
randomly generate $\{\hat\state_i\}$ using an accept-reject algorithm similar to the one in
\cite{Djeridane2006}, which is described in Alg. \ref{alg:alg_exp_filter}. 

\begin{algorithm}
\caption{Exponential Filter}
\label{alg:alg_exp_filter}
\begin{algorithmic}[1]
  \STATE \textbf{Result}: $\keep$
  \STATE \textbf{Inputs}: $\state$, $\mathcal{R}$, $\lambda$
  \STATE compute $\state_{\text{proj}}$ via \eqref{eq:xproj}
  \STATE generate $\beta$ uniformly from $[0, \lambda]$\;
  \IF{$\state \in \mathcal{R}$}
    \STATE $\keep$ = true\;
  \ELSE
    \IF{$\beta \le \lambda e^{-\lambda \Vert x - x_{\textrm{proj}} \Vert_2}$ }
        \STATE $\keep = $ true\;
    \ELSE
        \STATE $\keep = $ false\;
    \ENDIF
  \ENDIF
\end{algorithmic}
\end{algorithm}

\begin{algorithm}
\caption{Length Filter}
\label{alg:alg_length_filter}
\begin{algorithmic}[1]
  \STATE \textbf{Result}: $\mathcal{\trainset}$
  \STATE \textbf{Inputs}: $\trainset$, $\lambda_C$, $\hat{C}(\trainset)$, $D$
  \STATE generate $\beta$ uniformly from $[0, \lambda]$\;
  \FOR {$\state_i \in \trainset$}
  	\STATE $\costset \leftarrow \emptyset$\;
  	\FOR {$\state_j \ne \state_i \in \trainset: ||\state_i - \state_j||_2 \le D$}
  		\IF{$\beta > \lambda_C e^{-\lambda_C \hat{C}(\state_j)}$ }
    		\STATE $\costset \leftarrow \costset \cup x_j $\;
	  	\ENDIF
	\ENDFOR
	\STATE $\trainset \leftarrow \trainset \setminus \costset$ \;
  \ENDFOR
\end{algorithmic}
\end{algorithm}

Alg.~\ref{alg:alg_exp_filter} takes a state $\state$, an accept region $\mathcal
R$, and a decay rate $\lambda$ as inputs. To use Alg.~\ref{alg:alg_exp_filter}, we first compute $\state_\text{proj}$, the Euclidean projection of the state $\state$ onto the set $\mathcal{R}$ as follows:

\begin{equation}
\label{eq:xproj}
  x_\text{proj} = \arg\min_{x'} \Vert x' - x \Vert_2 : x' \in \mathcal{R}
\end{equation}

Using Alg.~\ref{alg:alg_exp_filter}, we generate two training sets: one large
dataset $\dataset_1$ and one small dataset $\dataset_2$. The large
data set $\dataset_1$, used for supervised training of the plant layer (the
weights $W_X$), is generated with a large accept region with a large number of
accepted points. The smaller dataset $\dataset_2$, used to initialize the
full RNN after the plant layer has been warmed up with $\dataset_1$, is generated with a smaller accept region and a relatively small number of accepted points.

Once the RNN's plant layer is warmed up with $\dataset_1$ and the full network
has been trained on $\dataset_2$, the RNN is ready to be queried, setting the
dynamic training loop into motion.

\subsubsection{Neural Network Query} \label{subsubsec:query}

At the start of every training loop, the RNN is used to predict controls,
$\{\hat\ctrl_i(\cdot)\}_{i=1}^{N}$, from a set of states $ \{\bar\state_i\}_{i=1}^{N}$.

For our training loop's first query, we uniformly sample $\emax$ states within a
distance of $\epsilon$ to $\bar\state$ to produce a set of states denoted
$\trainset_{\epsilon} = \{\bar\state_i\}_{i=1}^{\emax}$. Then, for all of the
network queries, we query the RNN with the same $\trainset_{\epsilon}$ to get
its current set of predictions, $\controlset_{\epsilon} =
\{\hat\ctrl_i(\cdot)\}_{i=1}^{\emax}$. By using a set of states,
$\trainset_{\epsilon}$, as opposed to using just a singular state, we capture
more local trajectories and relax the need for every trajectory to lead exactly
to a point location in the state space.


\subsubsection{Simulate Output and Feasibility} \label{subsubsec:sim}

In general, applying a control in $\controlset_\epsilon$ brings the backward system from
$\state_\targetset$ to some $\hat \state \not\in \trainset_{\epsilon}$. Thus, to
find $\controlset_{\epsilon}$'s true set of resultant states, we simply apply each
control in $\controlset\epsilon$ to $\dyn$ with $\state_\targetset$ as the
initial condition. This will yield $\controlset_{\epsilon}$'s true set of
resultant states, which we denote as $\trainset_{\epsilon}'$.

This key step is what gives us our feasibilty guarantee. Since all of the data
in our final output is a compilation of filtered data drawn from
$(\trainset_{\epsilon}', \controlset_{\epsilon})$ at each training cycle, we
know we have a dynamically feasible control for each state. In addition, since
the controls are feasible, they must also be either optimal or suboptimal.
\textbf{Therefore, when we compute values from our final set of controls using
    (\ref{eq:value_func}), these
values must be strictly conservative.}

\subsubsection{Filtering} \label{subsubsec:nnet_filt}




Often the RNN will predict a $\controlset_{\epsilon}$ that lead to states
far from our target. Since our method is intended to produce a value function
approximation local to a relevant region of the state space, we apply an
exponential accept-reject filter (Alg. \ref{alg:alg_exp_filter}) to
$\trainset'_{\epsilon}$, with accept region $\mathcal{R}_{\epsilon}$ and decay
rate $\lambda_{\epsilon}$, to stochastically remove states that are not nearby.
We let the remaining states and their corresponding controls be
$(\trainset_{\text{new}}, \controlset_{\text{new}})$.

We choose the input accept region $\mathcal{R}_{\epsilon}$ and decay rate
$\lambda_{\epsilon}$ provided to Alg.~\ref{alg:alg_exp_filter} such that the
filter will generously accept states that could lie near a feasible path from
$\bar\state$ to $\state_\targetset$. Though choosing a reasonable
$\mathcal{R}_{\epsilon}$ for general high-dimensional systems with complicated
dynamics before training could be difficult since we may be unable to
provide even a generous guess of where in the state space optimal trajectories
might lie, we can instead adjust the size or shape of
$\mathcal{R}_{\epsilon}$ until the region begins accepting predictions from the
RNN.

After finding $(\trainset_{\text{new}}, \controlset_{\text{new}})$, we can
update our full training set on which we will train our RNN, for the next training
iteration. We first improve our current training set, denoted as $(\trainset,
\controlset)$, by once again applying Alg.~\ref{alg:alg_exp_filter} with accept
region $\mathcal{R}$ and decay rate $\lambda$. Then, we apply a second filter to
all of our remaining data that favors controls with relatively low costs, this
is detailed in Alg.~\ref{alg:alg_length_filter} with inputs $\trainset$, the costs $\hat{C}(\trainset)$, and decay rate $\lambda_C$ and search radius $D$. We let the remaining data set
from filtering $(\trainset, \controlset)$ be called $(\trainset_{\text{old}},
\controlset_{\text{old}})$. Then, we compile our new training set for the next
training cycle as $(\trainset, \controlset) = (\{\trainset_{\text{old}},
\trainset_{\text{new}}\}, \{\controlset_{\text{old}},
\controlset_{\text{new}}\})$.

\subsection{Post-processing}\label{subsec:post_process}

In order to drive the system from $\bar x$ to $x_\targetset$, the value function, and in particular its gradient, is necessary at points between $\bar\state$ and $\state_\targetset$ along a dynamically feasible trajectory. Fortunately, this information can be computed from $\trainset$ and $\controlset$. Specifically, $\hat\state_i \in \trainset$ and $\hat\ctrl_i(\cdot) \in \controlset$ produce a trajectory $\traj^\dyn_i(0; \hat\state_i, t, \hat\ctrl_i(\cdot)), t \in [\hat T, 0]$. From the trajectories, we can obtain $M_i$ states on the trajectory by discretizing the time $t$ into $M_i$ points. We denote these states $\state_{(i,j)}$, where the index $i$ comes from the index of $\hat\state_i \in \trainset$, and the index $j \in \{0,\ldots,M_i-1\}$ indicates that the state is computed from the $j$th time point on the trajectory $\traj^\dyn_i$. Mathematically, $\hat\state_{(i,j)}$ is given as follows:

\begin{equation}
\label{eq:traj_trace}
\begin{aligned}
	\hat\state_{(i,j)} = \traj_i^\dyn(0; \hat x_i, t_j, \hat u_i (\cdot)), \\ 
  t_j = - \frac{j \hat T_i}{M_i-1}, j = 0, 1, \ldots, M_i-1
\end{aligned}
\end{equation}

Once we have explicitly added data along the trajectories from our dataset, we
now have a dataset that spans the local state space around and between
$\bar\state$ and $\state_{\targetset}$. To get our value function approximation
across our dataset, we can simply use (\ref{eq:value_func}). Explicitly, for
state-control pair, $(\hat x_i, \hat u_i(\cdot)) \in \trainset \times \controlset$, and $t_j$,
we have the approximate value function at states $x_{(i,j))}$ and times $t_j$:

\begin{equation}
\label{eq:approx_value}
\begin{aligned}
\hat V(x_{(i,j)}, t_j) = \int_{t_j}^{0} C(\hat{u}_i(t)) dt
\end{aligned}
\end{equation}




\section{Controller Synthesis Phase}\label{sec:post}

After we obtain our value function approximation, we can synthesize control to
drive our system from $\bar\state$ to $\state_{\targetset}$ using
\eqref{eq:opt_control} with the appropriate gradient components of the value
function. However, since our value function approximation is irregular and based
in a point set, we will first define a special computation for obtaining the
gradient.

To compute $(\nabla V)_i$, the $i$th component of the gradient at a given state $\state$, we search above and below in the $\hat{i}$ direction for states in $\trainset$ within a hyper-cylinder of tunable radius $r$. We define the closest point within the hyper-cylinder above $\state$ as $\state_a$ and below as $\state_b$. If some $\state_a$ and $\state_b$ exist, we compute the gradient at $\state$ as: 

\begin{equation}
\label{eq:grad_compute}
		(\nabla V(\state))_i = \frac{V(\state_a) - V(\state_b)}{(\state_a)_i - \state_b)_i}
\end{equation}

If we have multiple states above but none below, we approximate the gradient using (\ref{eq:grad_compute}), with $\state_a$ being the closest state above and $\state_b$ being the second closest state above. A similar procedure holds if we have multiple states below but none above.

In general, we can use a finite element method to compute gradient values for a
non-regular grid, which involves using shape functions as basis functions to interpolate the gradient values between nodes.

\section{Dubins Car Example}\label{sec:dubins_ex}

\begin{figure}[!t]
    \centering
    \includegraphics[width=0.9\columnwidth, trim = {0cm, 0cm, 0cm, .77cm},
    clip]{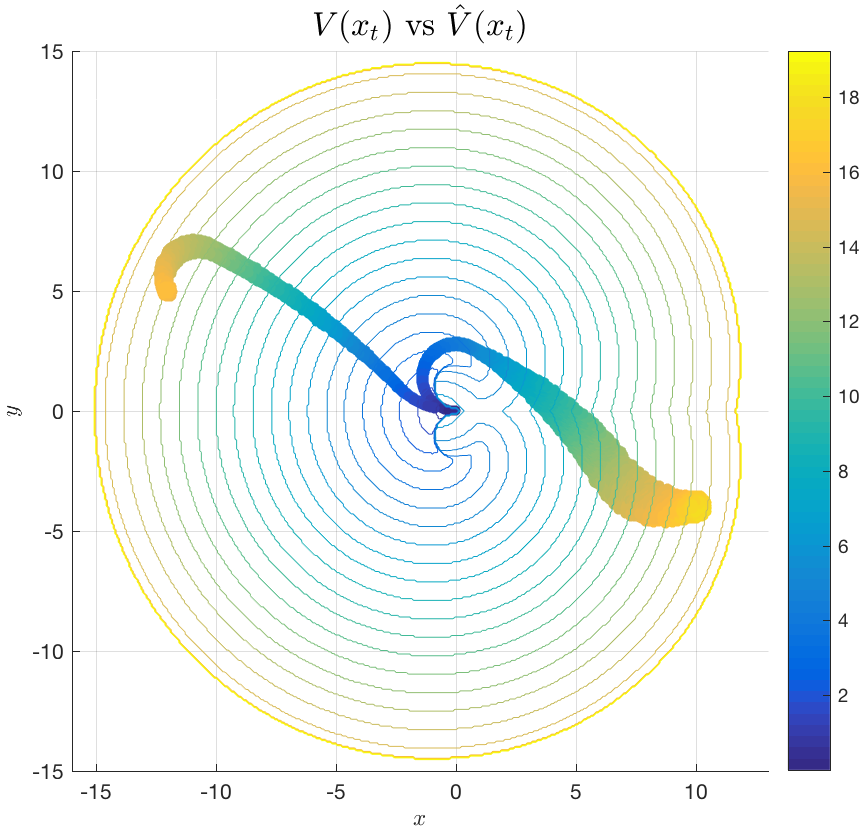}
    \caption{Comparison of the true value function $V(\state, t_0^*)$ computed over a large portion of the state space, and the approximate value function from our NN-based approach $\hat V(\state_{(i,j)}, t_j)$ on two different corridors containing the initial state $\bar\state$, the target set $\state_\targetset$, and a dynamically feasible trajectory. The contours are level sets of $V(\state, t_0^*)$. Two different corridors in which $\hat V$ is computed resulting from two different values of $\bar x$ ($(10, -4, -3), (-12, 5, 2)$) are also plotted using the same colormap.} 
    \label{fig:corridors}
\end{figure}

\subsection{Vehicle Dynamics}

Consider the Dubins Car \cite{Dubins1957}, with state $\state = (\pos_x, \pos_y, \theta)$. $(\pos_x, \pos_y)$ are the $x$ and $y$ positions of the vehicle, and $\theta$ is the heading of the vehicle. The system dynamics, assuming unit longitudinal speed, are

\begin{equation}
    \begin{aligned}
       \dot \pos_x = \cos \theta,& \ 
       \dot \pos_y = \sin \theta \\
       \dot \theta &= u, \qquad |\ctrl| \le 1
    \end{aligned}
\end{equation}

The control of the Dubins car is denoted $\ctrl$, and is constrained to lie in
the interval $[-1, 1]$, the interpretation of which is that the vehicle has a
maximum turn rate of $1$ rad/s. In addition, we only accrue cost on our control
with the duration of the control. Formally, this means that $C(u(t)) = 1$. We choose the Dubins car to illustrate our method because of the simple structure of the optimal controls. In addition, since the model is only 3D, we are able to verify our results by comparing them to the those obtained via HJ reachability.

For our example, we choose many different initial states $\bar \state$ for the system. The target state $\state_\targetset$ is chosen as the origin.

\subsection{Control Primitives}
In \cite{Dubins1957}, Dubins shows that all optimal trajectories of this
system utilize controls that represent going straight or turning maximally
left or right. Thus, the set of controls that are valid for generating optimal
trajectories can be reduced down to three motion primitives, \{`L', `S', `R'\},
encoding the controls $\ctrl = 1$ (max left), $\ctrl = 0$ (straight), and
$\ctrl=-1$ (max right) respectively. Though \cite{Dubins1957} additionally
provides an algebraic solution to the optimal control problem, we purposefully
do not leverage this result, as many interesting systems do not have such a
simple method of deriving optimal control. Instead, we use five motion primitives, encoding the controls $\ctrl = 2$, $\ctrl = 1$, $\ctrl = 0$, $\ctrl = -1$, $\ctrl = -2$. 


Following our notation in Section \ref{sec:training_phase}, we write a $n$
length control
sequence as $\{u^1, u^2, \ldots, u^n\}, \{\ctime^1,\ctime^2, \ldots, \ctime^n\}$ where
$u^i$ denotes the $i$th control primitive and $\tau^i$ denotes the duration of
$i$th control primitive.

\subsection{Neural Network}

Using the RNN architecture described in Section \ref{subsec:nnet}, we let
$N=3$, since we only need at most three control primitives. Since the controls
and dynamics of Dubins car are simple, we have chosen the number of neurons in
layers $P, D, U, X$ to be $[10, 10, 6, 75]$, respectively. The NN is trained
with the training functionality of the MATLAB Neural Network ToolBox 2016a. The
training function used for the full NN is resilient back-propagation and the
performance function used is mean squared error.

\subsection{Dynamic Training}



For $\dataset_1$ and $\dataset_2$, we choose each $\ctrl^i$ from the three possible
values $\{-1, 0, 1\}$. The controls for $\dataset_1$ have durations, $\tau^i$,
uniformly sampled from $[0, \bar T]$ where $\bar T = 100$. $\dataset_2$'s control
durations are also uniformly sampled in the same manner, but with $\bar T=2\pi$.

For the dynamic training parameters, our query set $\trainset_{\epsilon}$ is
generated with $\epsilon = 1$ and $M_\epsilon = 500$.

\subsubsection{Filtering Algorithms}
\label{sec:result_filter}


Throughout training, the exponential filtering process uses two accept regions. In early iterations, $\mathcal R = \mathcal R_c$ is defined as the cone
of minimum size that contains $\trainset_{\epsilon}$, with the tip of the cone located
at $\bar x$. The choice of using a conical filter to guide the neural net at first is based on the hypothesis
that a trajectory taking the system from $\bar\state$ to $x_\targetset$ is
likely to stay in the cone $\mathcal R_c$. In later iterations, $\mathcal R = \mathcal R_s$ is chosen as $\trainset_\epsilon$. This spherical filter, centered at $x_\targetset$, helps to more finely guide the neural network to $x_\targetset$. 


We also apply length filtering with a small distance parameter $D = 0.5$, since we would like 
to be comparing distance costs only between trajectories with similar end points.

\subsubsection{Filter Decay Rate: Setting and Timing}
Although $\mathcal R_\text{C}$ and $\mathcal R_\text{S}$ are chosen before the dynamic training process, the filtering of the training set $\trainset$ can still be adjusted while training. This is done by varying the parameters $\lambda_\text{C}$ and $\lambda_\text{S}$. In early training iterations, we want to decrease $\lambda_\text{C}$ slightly to ensure that we are not filtering out states needed for the NN to explore the state space. Once the NN has gained a better understanding of how to reach $\bar\state$, we increase $\lambda_\text{C}$ and $\lambda_\text{S}$ slightly to further encourage the NN to drive states near $\bar\state$. When the dataset is mostly near $\bar\state$, we increase $\lambda_\text{S}$ and $\lambda_\text{S}$ significantly. The decay rate for the length filter, $\lambda_C$, is constant over iterations.

\subsection{Dubins Car Results}
\label{sec:results}
\subsubsection{Training Process}

\begin{figure}[!t]
  \centering
  \begin{subfigure}{4cm}
    \raggedleft
    \includegraphics[width=4cm]{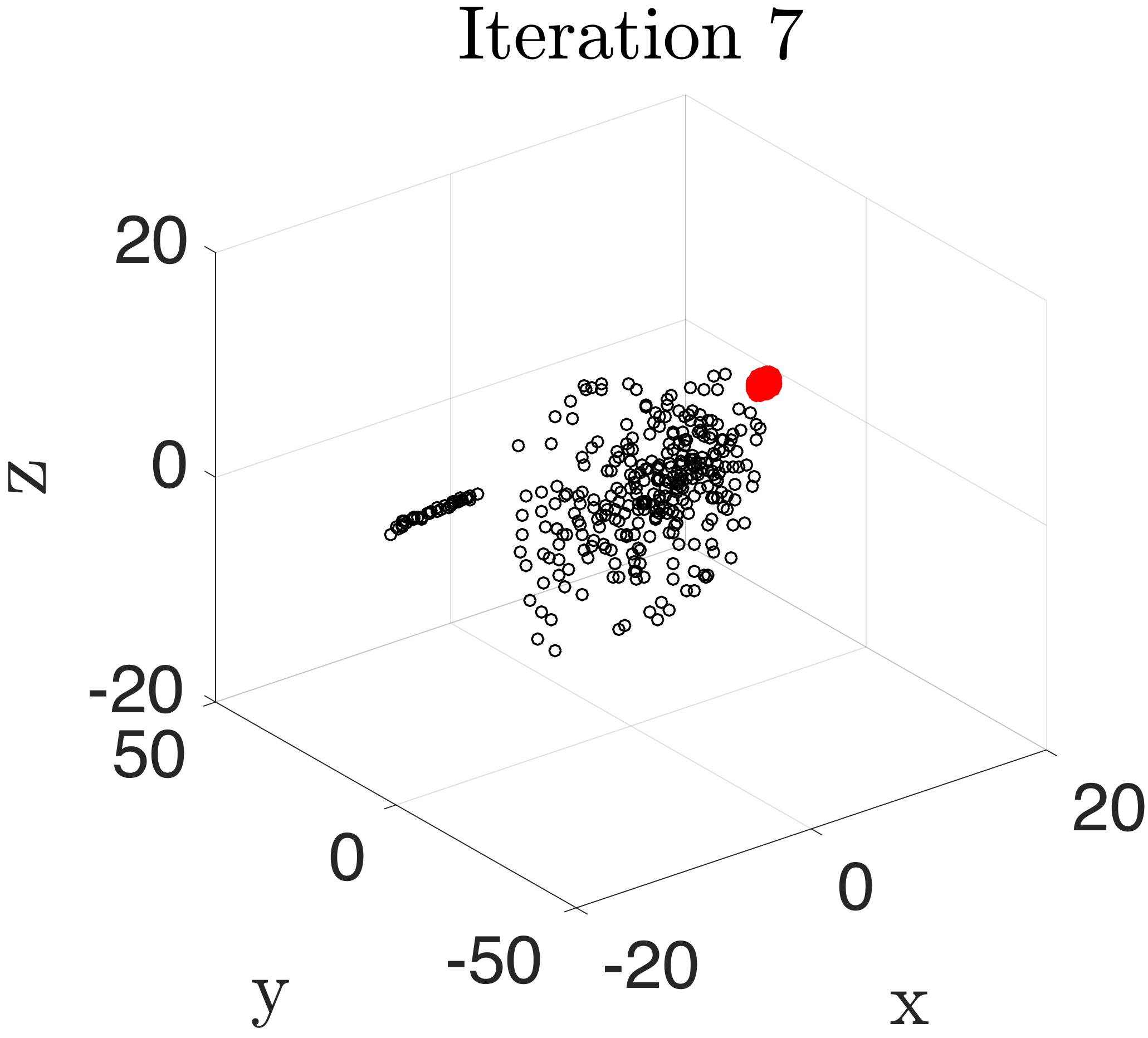}
    \caption{\label{fig:training_a}}
  \end{subfigure}
  \hfill
  \begin{subfigure}{4cm}
    \raggedright
    \includegraphics[width=4cm]{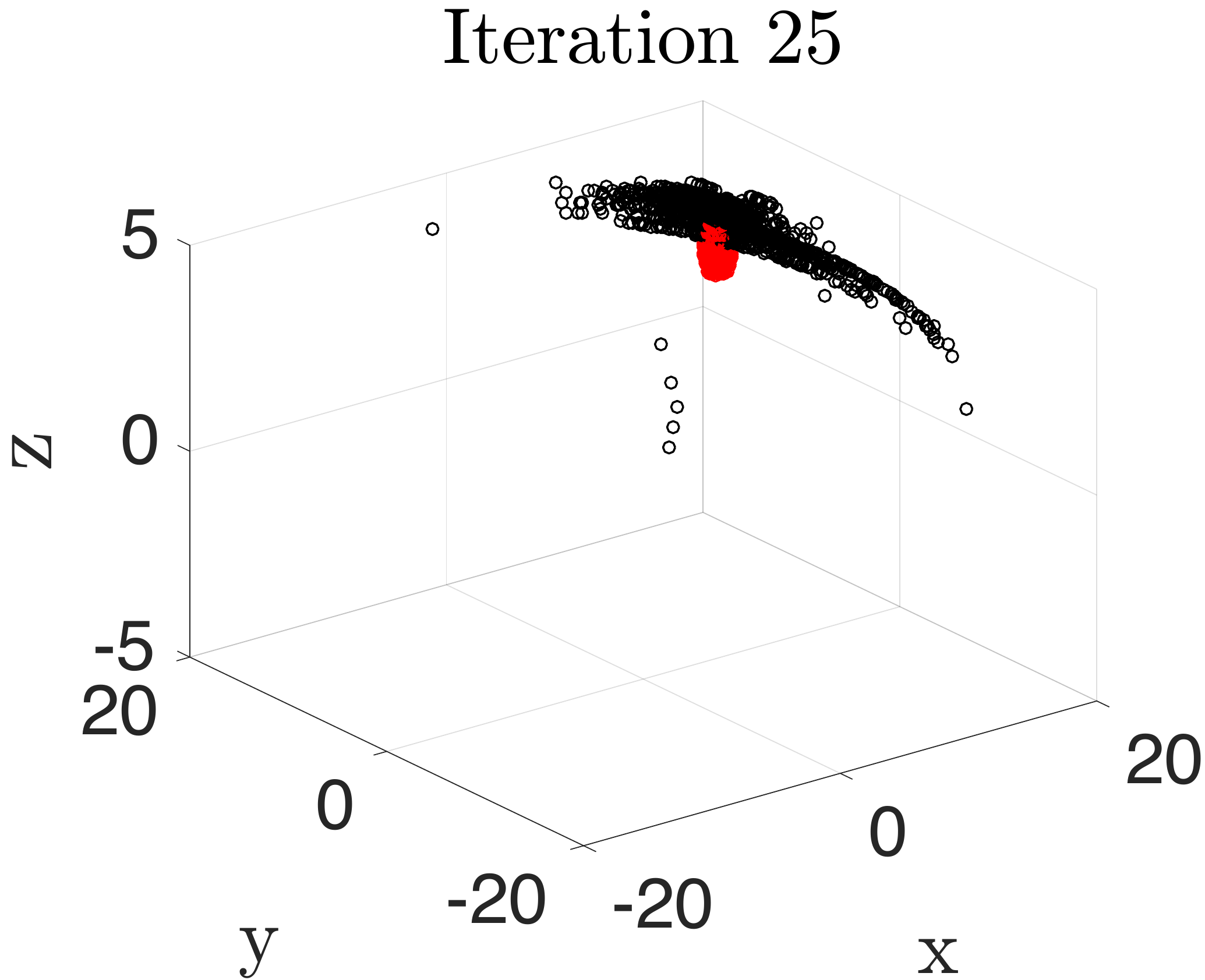}
    \caption{\label{fig:training_b}}
  \end{subfigure}
  \hfill
  \begin{subfigure}{8cm}
    \centering
    \includegraphics[width=5cm]{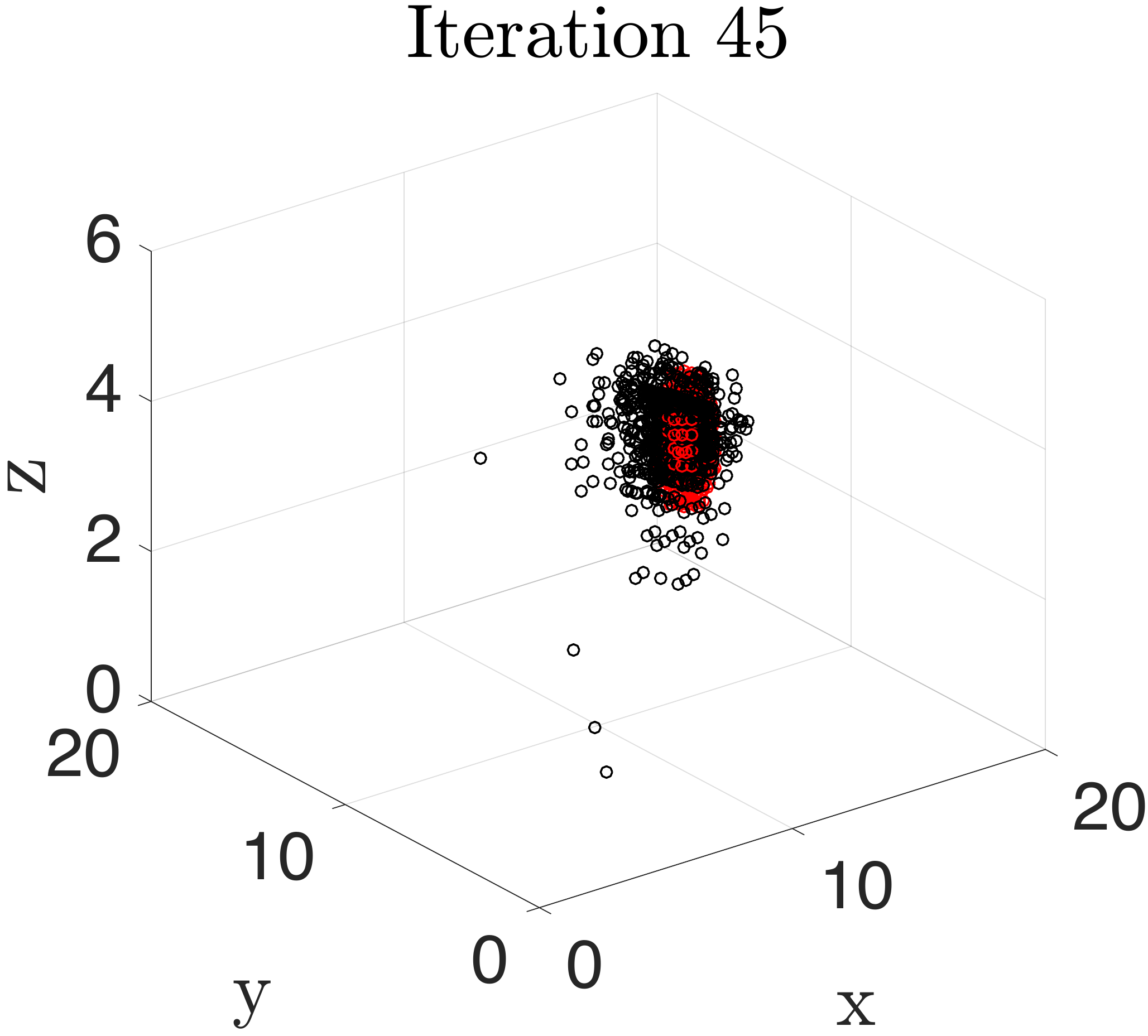}
    \caption{\label{fig:training_c}}
  \end{subfigure}
  \caption{Evolution of $\trainset$ over many iterations.}
\end{figure}

In Fig. \ref{fig:training_a}, \ref{fig:training_b}, \ref{fig:training_c}, the
process by which the training set $\trainset$ iteratively changes from the
initial training set $\dataset_2$ to encompassing $\trainset_{\epsilon}$ 
is shown. Here, the red states represent the set $\trainset_{\epsilon}$ and the
black states represent the current states in $\trainset$. In the early
iterations (Fig. \ref{fig:training_a}), the NN explores outward from the initial
training set, frequently making mistakes, resulting in the states in $\trainset$
being very far away from $\bar\state$.  As the iteration number increases, the
trajectory ambiguous training set $\dataset_2$ is gradually cut down, and
eventually the NN begins to predict controls $\hat\ctrl_i(\cdot)$ that drive to
states $\hat\state_i$ in an arc that heavily intersects $\trainset_{\epsilon}$.
This can be seen in Fig. \ref{fig:training_b}. By the end of the training
process, the $\mathcal R_O$ conical target filter prunes the states outside of
the $\trainset_{\epsilon}$. This can be seen in Fig. \ref{fig:training_c}.

\begin{figure}[!b]
  \centering
  \begin{subfigure}{.49\columnwidth}
    \includegraphics[width=\columnwidth]{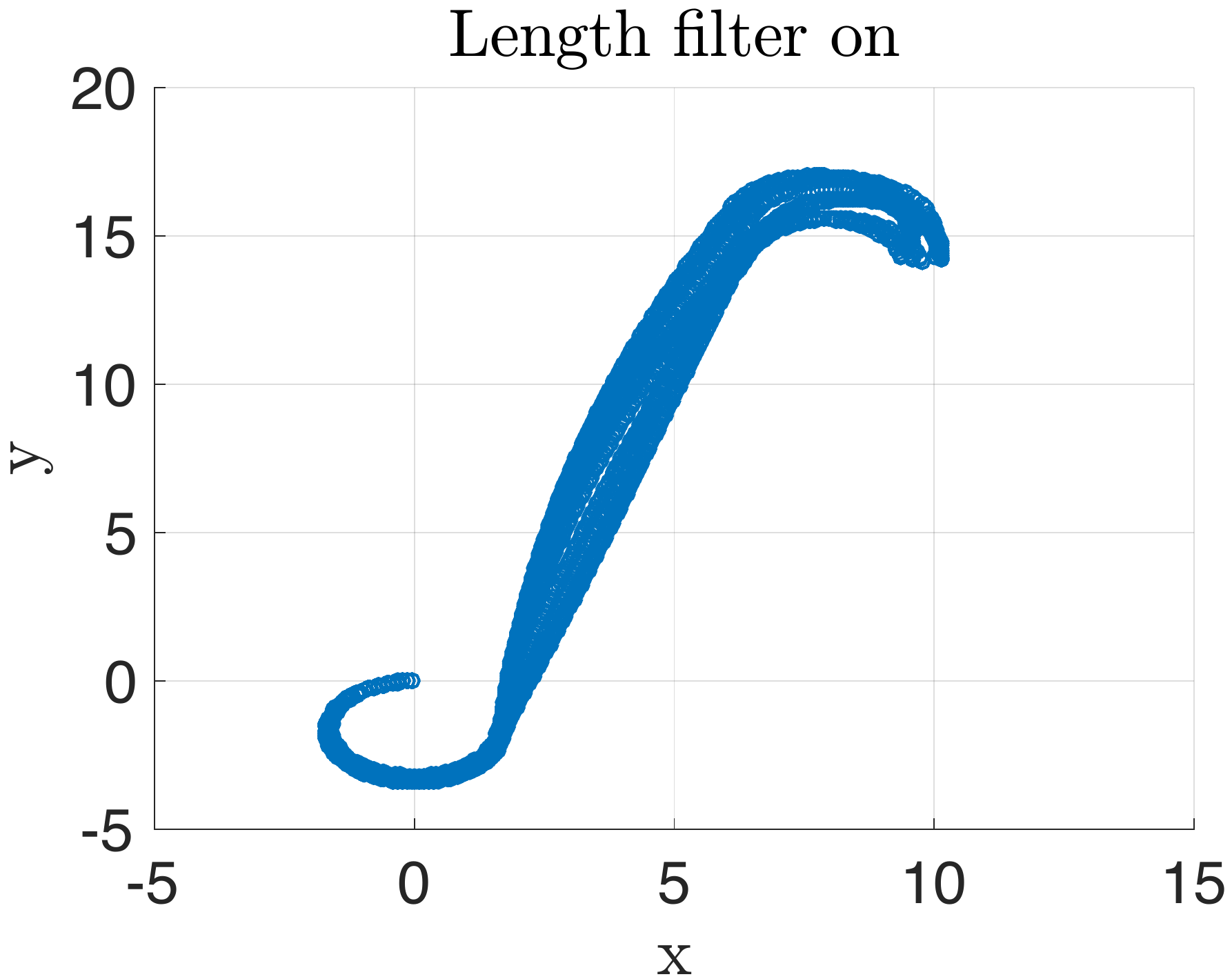}
	\caption{\label{fig:lengthfilter}}
  \end{subfigure}
  \hfill
  \begin{subfigure}{.49\columnwidth}
    \includegraphics[width=\columnwidth]{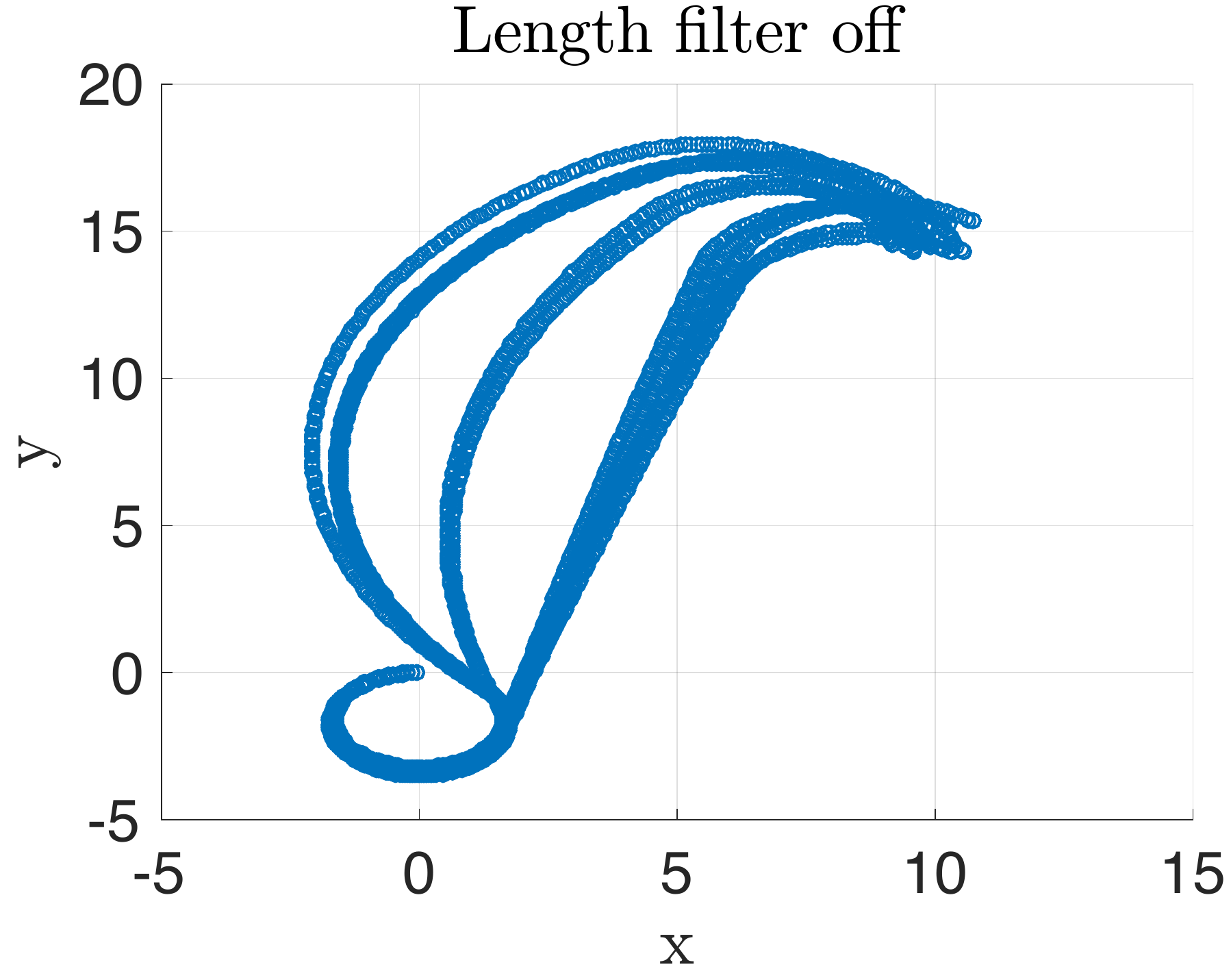}
    \caption{\label{fig:nolengthfilter}}
  \end{subfigure}
  \caption{Effect of the length filter on quality of $\trainset$. (\subref{fig:lengthfilter}) shows $\trainset$ when the filter is on, and (\subref{fig:nolengthfilter}) shows $\trainset$ when it is off.}
\end{figure}

The length filter also enables our method to be robust to suboptimal training data.
If we provide the neural net with a mixture of optimal and suboptimal training data, the length filter improves the quality of $\trainset$ by removing many states generated by using suboptimal control (Figure \ref{fig:lengthfilter}), compared to without the length filter (Figure \ref{fig:nolengthfilter}).

\subsubsection{Value Function Comparison}

Using level set methods \cite{Mitchell05}, we computed $V(x, t)$, and compared the true value function, $V(\bar\state, t_0^*)$ and the approximate value function, $\hat V(\bar\state, \hat T)$ computed for several states in Table~\ref{table_times}. $t_0^*$ denotes the time component of the optimal solution of \eqref{eq:value_func}.


\begin{table}
  \centering
  \begin{tabular}{|P{1.4cm}||P{2.3cm}|P{2.3cm}|}
    \hline
    State $\bar\state$ & NN Cost $\hat{V}(\bar\state, \hat{T})$          & True Cost $V(\bar\state, t_0^*)$ \\ \hline
 $(-12, 5, 2)$   & $26.51$   &   $14.84$\\
 $(-10, 0, 0)$   & $10.23$  &   $10.00$\\
 $(1, 1, 6)$   & $9.93$   &   $7.40$\\
 $(10, -4, -3)$   & $16.20$  &   $13.36$\\
  \hline
  \end{tabular}
  \caption{Trajectory values (seconds)}
  \label{table_times}
\end{table}

\subsubsection{Computation Time}

Synthesizing control using our NN-based approach allows for large time
complexity improvements in comparison to using level set methods. On a 2012
MacBook Pro laptop, data generation requires approximately 3 minutes,
and controller synthesis from this data and simulation requires
2 minutes on average. Since the region of the state space we are considering is quite large, and the target set is quite small (a singleton), the level set methods approach is intractable on this laptop, and requires 4 days on a desktop computer with a Core i7-5820K processor and 128 GB of RAM. 

There are also large spatial savings by using the NN. For example, $\trainset$
and $\hat V$ for one particular corridor computed between $(10, -1, -3)$ and $(0, 0, 0)$ requires only 179 MB, while a reachable set computed over that horizon on a very low resolution grid requires approximately 7 GB.

As can be seen from this and the previous sections, using level set methods not only is more time-consuming compared to using our NN-based approach, but also does not guarantee a more shorter trajectory due to discretization error.

\begin{figure}[!t]
    \centering
    \includegraphics[width=0.9\columnwidth]{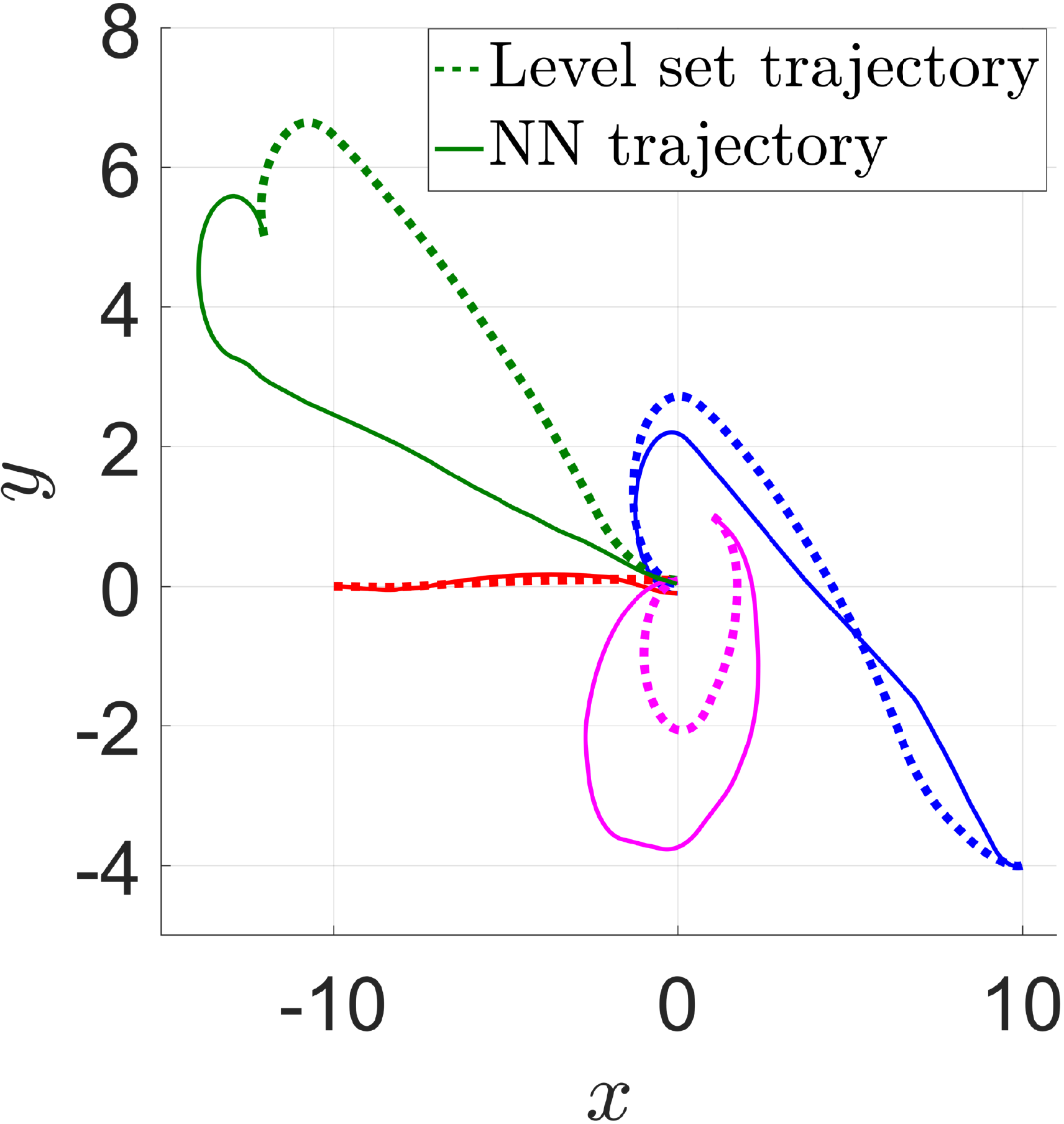}
    \caption{Trajectories generated using level set methods (dashed) and using our NN-based method (solid). Each color corresponds to a different initial state $\bar x$.}
    \label{fig:fig_traj}
\end{figure}

\section{Conclusions and Future Work}\label{sec:conclusion}
Our NN-based grid-free method computes an upper bound of the optimal value
function in a region of the state space that contains the initial state, the
target set, and a feasible trajectory. By combining the strengths of dynamic
programming-based and machine learning-based approaches, we greatly alleviate
the curse of dimensionality while maintaining a desired direction of
conservatism, effectively avoiding the shortcomings of both types of approaches.


Using a numerical example, we demonstrate that our approach can successfully
generate a value function approximation in multiple test cases for the Dubins
car. We are even able to approximate value function values in regions that are
very far from the target set, a very computationally expensive task for dynamic
programming-based approaches. Our approximate value function is able to drive
the Dubins car from many different initial conditions to the target set.

Although our current results are promising, much more investigation is still
needed to make our approach more practical and applicable to more scenarios. For
example, better intuition for the choice of accept regions in the filtering
process is needed to extend our approach to other systems. We currently plan to
investigate applying our method to the 6D engine-out plane \cite{Adler2012} as
well as a 12D quadrotor model. In addition to path planning, we also hope to
extend our theory to provide safety guarantees and robustness against
disturbances. Such extensions are non-trivial due to the different roles that
the control and disturbance inputs play in the system dynamics.


\bibliographystyle{IEEEtran}
\bibliography{references}

\end{document}